\documentclass[11pt]{article}

\usepackage[final]{acl}

\usepackage{times}
\usepackage{latexsym}
\usepackage{booktabs}
\usepackage[T1]{fontenc}

\usepackage[utf8]{inputenc}

\usepackage{microtype}

\usepackage{inconsolata}
\usepackage{booktabs}
\usepackage{xcolor}
\usepackage{colortbl}
\usepackage{graphicx}
\usepackage{amsmath}
\usepackage{algorithm}
\usepackage{algpseudocode}
\usepackage{amssymb}
\usepackage{bm}

\newcommand{\hl}[1]{\colorbox{yellow!40}{#1}}
\title{ SOMP: Scalable Gradient Inversion for Large Language Models via Subspace-Guided Orthogonal Matching Pursuit
}

\author{Yibo Li \\
  Politecnico di Milano / Italy \\
  \texttt{yibo2.li@mail.polimi.it} \\\And
  Qiongxiu Li \\
  Aalborg University \\
  \texttt{qili@es.aau.dk} \\}

\begin{document}
\maketitle
\begin{abstract}

Gradient inversion attacks reveal that private training text can be reconstructed from shared gradients, posing a privacy risk to large language models (LLMs). While prior methods perform well in small-batch settings, scaling to larger batch sizes and longer sequences remains challenging due to severe signal mixing, high computational cost, and degraded fidelity.
We present \textbf{SOMP} (Subspace-Guided Orthogonal Matching Pursuit), a scalable gradient inversion framework that casts text recovery from aggregated gradients as a sparse signal recovery problem. Our key insight is that aggregated transformer gradients retain exploitable head-wise geometric structure together with sample-level sparsity. SOMP leverages these properties to progressively narrow the search space and disentangle mixed signals without exhaustive search. Experiments across multiple LLM families, model scales, and five languages show that SOMP consistently outperforms prior methods in the aggregated-gradient regime.For long sequences at batch size $B=16$, SOMP achieves substantially higher reconstruction fidelity than strong baselines, while remaining computationally competitive. Even under extreme aggregation (up to $B=128$), SOMP still recovers meaningful text, suggesting that privacy leakage can persist in regimes where prior attacks become much less effective.
\end{abstract}

\section{Introduction}

The reliance of LLMs on massive, often sensitive datasets raises critical privacy concerns, particularly in collaborative training scenarios across multiple institutions.
To address these privacy concerns across sensitive domains (e.g., healthcare~\cite{TEO2024101419}, law~\cite{zhang-etal-2023-fedlegal}, and finance~\cite{10384119}), Federated Learning (FL) protects raw data by sharing only aggregated gradient updates~\cite{zhang2024buildingfederatedgptfederated,yao2025federatedlargelanguagemodels}. Consequently, relying on large local batch sizes is typically deployed to obfuscate individual samples, creating a strong expectation of data privacy.

Despite its appeal, federated learning does not provide inherent privacy guarantees~\cite{li2024perfectgradientinversionfederated}. Early gradient inversion attacks showed that an honest-but-curious server can reconstruct private training data directly from shared gradients~\cite{zhu2019deepleakagegradients}. Although these attacks were first studied in continuous image domains~\cite{li2022auditingprivacydefensesfederated}, they now pose a direct threat to LLMs, where discrete textual inputs can also be recovered. In particular, methods such as \emph{DAGER} demonstrate that entire text batches can be exactly reconstructed from aggregated transformer updates. These results directly challenge the common assumption~\cite{mcmahan2023communicationefficientlearningdeepnetworks} that gradient aggregation alone is sufficient to obfuscate individual samples.

However, existing text-based gradient inversion methods face serious scalability challenges. Early approaches rely heavily on external language models and are therefore limited to short sequences or lightly aggregated settings. More recent methods extend to larger setups, but introduce new bottlenecks: hybrid optimization methods such as GRAB~\cite{10.1145/3658644.3690292} suffer from cross-sample interference, while exact recovery methods such as DAGER~\cite{petrov2024dagerexactgradientinversion} incur prohibitive costs under longer sequences and large batchsize.  As a result, prior methods become brittle precisely in the long-sequence, multi-sample regimes that are most relevant in practice.

To address this gap, we propose \textbf{SOMP},  a structured framework for multi-sample text gradient inversion under aggregated gradients. We study the standard honest-but-curious server setting, where the attacker observes an aggregated gradient update. Our key insight is that aggregated transformer gradients are not arbitrary mixtures: they retain useful head-wise geometric structure induced by multi-head attention, together with sample-level sparsity. Building on this observation, SOMP reformulates inversion as a structured reconstruction problem in gradient space, enabling more effective disentanglement without exhaustive token-level verification. Our goal is not to claim uniform gains in every regime, but to improve reconstruction quality and scaling behavior precisely where prior attacks degrade most: long-sequence, large batchsize settings.

\noindent$\bullet$ \textbf{Head-aware geometric guidance for text gradient inversion.}
We show that aggregated transformer gradients retain useful head-wise geometric structure, and exploit this structure to guide candidate filtering and decoding under mixed gradients.

\noindent$\bullet$ \textbf{A sparse reconstruction formulation for multi-sample inversion.}
We reformulate exact recovery from exhaustive token-level search as sparse signal recovery in gradient space, substantially mitigating the scalability bottlenecks of prior methods and enabling effective disentanglement of cross-sample interference.

\noindent$\bullet$ \textbf{Robust and scalable inversion under gradient aggregation.}
Across multiple transformer families and aggregated-gradient regimes, SOMP yields stronger reconstruction quality than prior text-based attacks, with its largest gains appearing in long-sequence and larger-batch settings.

\section{Related Work}

\paragraph{Gradient inversion and data leakage attacks.}
The problem of reconstructing private data from shared gradients, often termed \emph{gradient inversion} or \emph{gradient leakage}, was first explored~\cite{zhu2019deepleakagegradients} and has since drawn extensive attention in the context of federated learning. 
Early approaches mainly targeted image data and formulated the recovery as a continuous optimization problem~\cite{zhu2019deepleakagegradients, chen2025gradientinversiontranscriptleveraging, huang2021evaluatinggradientinversionattacks}. More recent works such as \textbf{SPEAR}~\cite{dimitrov2024spearexactgradientinversionbatches} further improved efficiency by exploiting the inherent low-rank and sparsity structures of neural gradients, enabling exact reconstruction of large image batches. 
However, these methods rely on continuous optimization and cannot be directly transferred to text, where inputs are discrete and tokenization introduces non-differentiable constraints.~\cite{geng2022generaldeepleakagefederated}

\paragraph{Gradient inversion for textual data.}
Reconstructing text from gradients is substantially more challenging than
recovering images due to the discrete token space and the combinatorial
structure of language.
Early approaches such as \textbf{LAMP}~\cite{balunović2022lampextractingtextgradients}
and \textbf{GRAB}~\cite{10.1145/3658644.3690292} incorporate language priors
or hybrid continuous--discrete optimization, but remain limited in
scalability under aggregated or long-sequence settings due to the exponentially growing search space and severe cross-token interference.
\textbf{DAGER}~\cite{petrov2024dagerexactgradientinversion}
exploits the low-rank structure of self-attention gradients to enable
exact reconstruction in transformer architectures.
However, DAGER suffers from high computational cost due to layer-wise
exhaustive search and becomes unstable in high-token regimes, where
aggregating multiple inputs or long sequences causes the
attention-gradient matrix to approach full rank, degrading reconstruction
accuracy.

\section{Background}
\label{background}

In this section, we introduce the problem setup and summarize the structural properties of gradients in transformer-based models that serve as fundamentals of our method.

\subsection{Problem Setup}

We consider LLMs parameterized by $\bm{\theta}$ and trained under the \emph{FedSGD} protocol. Each client holds a local mini-batch of $B$ training samples $\{(x_j, y_j)\}_{j=1}^{B}$ and transmits only the aggregated gradient to the server. Given a loss function $\mathcal{L}$, the server observes
\begin{equation}
\bm{g}_{\text{mix}}
=
\frac{1}{B}
\sum_{j=1}^{B}
\bm{\nabla_\theta \mathcal{L}}\bigl(f_\theta(x_j), y_j\bigr),
\label{eq:mixedGradient}
\end{equation}
without access to individual samples or per-sample gradients.

We consider a transformer with context length $p$, hidden dimension $d$, and $H$ attention heads. Given an input sequence $x = [x_1, \ldots, x_p]$, the corresponding token embeddings form the input representation $\bm{Z}_0 \in \mathbb{R}^{p \times d}$.

\subsection{Structural Properties}

Gradients in transformer-based models exhibit several structural properties that are useful for multi-sample inversion. First, as noted in prior work~\cite{petrov2024dagerexactgradientinversion}, gradients of linear layers inherit a matrix-product form. For a linear layer $Y = XW + (b|\cdots|b)^\top$, the weight gradient satisfies
\begin{equation}
\bm{\nabla_W \mathcal{L}} = X^\top \nabla_Y \mathcal{L},
\label{eq:lowrank}
\end{equation}
which induces a low-rank structure in the resulting gradient matrix.

Second, multi-head attention induces a natural head-wise decomposition in gradient space. Let $(\cdot)_l$ denote the $l$-th transformer layer, $(\cdot)_{Q,K,V}$ denote the query, key, and value projections, and $(\cdot)^{(h)}$ index the $h$-th attention head. In particular, $\bm{W}_{1,Q}^{(h)}$ denotes the query projection matrix of head $h$ in the first transformer layer. For the first-layer query projection, the gradient of head $h$ can be written as
\begin{equation}
\bm{\nabla W}_{1,Q}^{(h)}
=
\bm{Z_0}^\top \bm{G}_Q^{(h)} \in \mathbb{R}^{d \times d_h},
\label{eq:queryGradientSlice}
\end{equation}
where $\bm{G}_Q^{(h)} \in \mathbb{R}^{p \times d_h}$ denotes the backpropagated signal for head $h$, and $d = \sum_{h=1}^{H} d_h$. The full query gradient is obtained by concatenating these head-wise slices.

Finally, these gradients exhibit informative sparsity patterns at the token level~\cite{dimitrov2024spearexactgradientinversionbatches}. In particular, head-aligned sparsity provides useful cues for filtering and scoring candidate tokens under aggregated gradients. A formal discussion is provided in Appendix~\ref{app:head_sparsity}.

Together, these properties suggest that aggregated gradients retain useful structure that can be exploited for multi-sample reconstruction.

\section{Methodology}
\label{sec:methodology}
\begin{figure*}[t]
\centering
\includegraphics[width=0.7\textwidth]{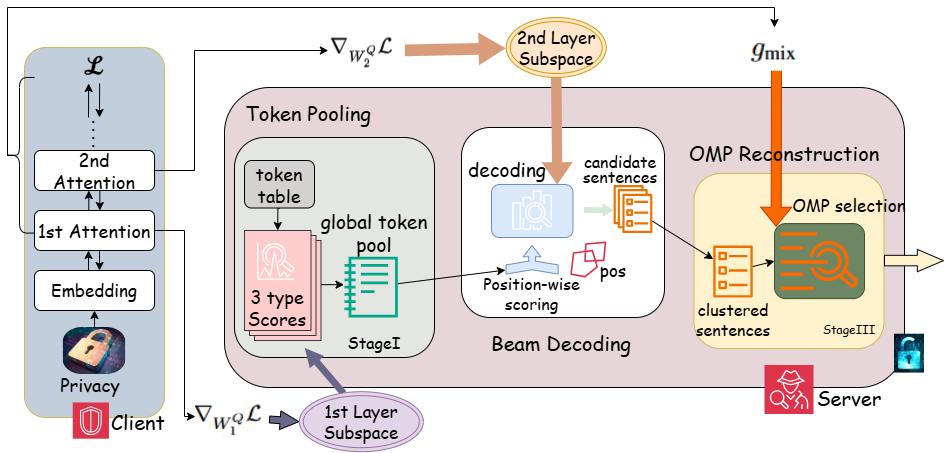}
\caption{Overview of SOMP. It consists of Stage~I head-structured token pooling, Stage~II geometry-driven diverse decoding, and Stage~III gradient-space sparse reconstruction.}
\label{fig:somp-overview}
\end{figure*}

We now proceed to introduce the proposed \textbf{SOMP}, a three-stage framework for reconstructing textual inputs from the aggregated gradient $\bm{g}_{\text{mix}}$. Unlike prior search-based methods, SOMP treats multi-sample inversion as a structured reconstruction problem and progressively narrows the search space from tokens to sentences and finally to sample gradients. Stage~I uses head-wise gradient slices to construct a compact token pool. Stage~II decodes candidate sentences from this pool using geometry-guided beam search with an LM prior. Stage~III then treats the decoded candidates as gradient atoms and selects a sparse subset whose gradients best reconstruct $\bm{g}_{\text{mix}}$. This progressive design separates coarse candidate filtering from final sample selection, making reconstruction more scalable under long sequences and large-batch aggregation. The overall pipeline is illustrated in Figure~\ref{fig:somp-overview}.

Unless otherwise specified, hyperparameters are fixed across experiments. A full list of hyperparameters is provided in Appendix~\ref{app:hyperparameters}.

\subsection{Stage I: Head-Structured Token Pooling}
\label{Stage:I}

Following Eq.~(\ref{eq:queryGradientSlice}), we decompose the first-layer query gradient into head-specific slices
$\{\bm{\nabla W}_{1,Q}^{(h)}\}_{h=1}^{H}$.
Each slice induces a head-specific subspace, which we use to evaluate token plausibility under the observed aggregated gradient. The goal of Stage~I is not to reconstruct complete sequences, but to build a compact, high-recall token pool that constrains the search space in Stage~II.

To rank candidate tokens, we combine three complementary signals: geometric alignment to the head-wise query subspaces, consistency across informative heads, and head-aware sparsity patterns in the FFN gradient.

Detailed pseudocode is given in Appendix~\ref{app:algorithms} (Algorithm~\ref{alg:stage1}).

\paragraph{Subspace-alignment score.}
For a candidate embedding $\bm{e}(v,pos)\in\mathbb{R}^d$, we first project it into the query space of head $h$:
\[
\bm{q}^{(h)}(v,pos)=\bm{e}(v,pos)\bm{W}^{(h)}_{1,Q}\in\mathbb{R}^{d_h}.
\]
Let $\bm{P}_{\mathcal{R}_Q^{(h)}}$ denote the orthogonal projector onto the subspace induced by $\bm{\nabla W}_{1,Q}^{(h)}$. We define the head-wise alignment score as
\begin{equation}
s_{\text{sub}}^{(h)}(v,pos)
=
\left\|
(\bm{I}-\bm{P}_{\mathcal{R}_Q^{(h)}})\bm{q}^{(h)}(v,pos)
\right\|_2,
\label{eq:MultiHeadsubspace_score}
\end{equation}
where smaller values indicate stronger geometric compatibility. We then aggregate these scores across the informative heads $\mathcal{H}_{\text{act}}$:
\begin{equation}
s_{\text{sub}}(v,pos)
=
\frac{1}{|\mathcal{H}_{\text{act}}|}
\sum_{h\in\mathcal{H}_{\text{act}}}
s_{\text{sub}}^{(h)}(v,pos).
\label{eq:subspace_score}
\end{equation}

\paragraph{Cross-head consistency score.}
Plausible tokens should align consistently across informative heads rather than match only a single head by chance~\cite{li2019informationaggregationmultiheadattention,voita-etal-2019-analyzing}. We therefore compute
\begin{equation}
s_{\text{cons}}(v,pos)
=
\mathrm{Std}_{h\in\mathcal{H}_{\text{act}}}
\bigl(s_{\text{sub}}^{(h)}(v,pos)\bigr),
\label{eq:score_consistency}
\end{equation}
where smaller values indicate more stable cross-head alignment.

\paragraph{Head-aware sparsity score.}
In addition to geometric alignment, we exploit token-level sparsity signals using head-aligned FFN blocks. For each block $h$, we compute
\begin{equation}
\bm{u}^{(h)}(v,pos)
=
\bm{e}(v,pos)^\top \bm{W}_{\mathrm{FFN}}^{(h)}
\in \mathbb{R}^{d_h},
\end{equation}
where $\bm{W}_{\mathrm{FFN}}^{(h)}\in\mathbb{R}^{d\times d_h}$ denotes the FFN gradient slice associated with block $h$. The corresponding block-wise sparsity score is
\begin{equation}
s^{(h)}(v,pos)
=
\frac{1}{d_h}
\sum_{m=1}^{d_h}
\mathbf{1}\!\left(
|u_m^{(h)}(v,pos)| \le \tau_m^{(h)}
\right),
\end{equation}
where $\tau_m^{(h)}$ is a per-dimension threshold estimated from the candidate set. Let $\mathcal{H}_{\text{top}}$ denote the top-$k$ blocks ranked by sparsity. We aggregate them as
\begin{equation}
s_{\text{sparse}}(v,pos)
=
\frac{1}{k}
\sum_{h\in\mathcal{H}_{\text{top}}}
s^{(h)}(v,pos).
\label{eq:score_sparity}
\end{equation}

\paragraph{Token ranking.}
We then combine the three scores into a single cost:
\begin{equation}
\begin{aligned}
s_{\text{total}}(v,pos)
&=
\lambda_1 s_{\text{sub}}(v,pos)
+\lambda_2 s_{\text{cons}}(v,pos) \\
&\quad
-\lambda_3 s_{\text{sparse}}(v,pos).
\end{aligned}
\end{equation}
where lower values indicate higher token plausibility. Tokens are ranked by minimizing $s_{\text{total}}$, and the top candidates are retained to form the token pool used in Stage~II.

\subsection{Stage II: Geometry-Driven Diverse Beam Decoding}
\label{Stage:II}

Stage~II generates candidate sentences from the token pool produced by Stage~I. To guide decoding, we construct head-wise query-gradient subspaces
$\{\mathcal{R}^{(h)}_{2,Q}\}$ from the second transformer layer, following prior observations that lower and middle layers provide more stable geometric directions for inversion.See Algorithm~\ref{alg:stage2} for the full decoding procedure.

At decoding step $t$, let $h_t^{(h)}$ denote the head-wise hidden representation for head $h$. We measure its mismatch to the recovered geometry by
\[
d_t^{(h)}
=
\left\|
(\bm{I}-\bm{P}_{\mathcal{R}^{(h)}_{2,Q}})
h_t^{(h)}
\right\|_2.
\]
We then aggregate these distances across informative heads:
\[
d_t^{\mathrm{geo}}
=
\frac{1}{|\mathcal{H}_{\text{act}}|}
\sum_{h\in\mathcal{H}_{\text{act}}}
d_t^{(h)}.
\]

To improve fluency, we combine the geometric score with an LM prior~\cite{devlin-etal-2019-bert,dathathri2020plugplaylanguagemodels}. Let $\tilde{h}_t$ denote the final-layer hidden state and let $v_{x_t}$ denote the embedding of candidate token $x_t$. We define
\begin{equation}
s_{\mathrm{LM}}=\langle \tilde{h}_t, v_{x_t} \rangle,
\end{equation}
and combine it with geometry as
\begin{equation}
d_t = d_t^{\mathrm{geo}} - \beta_{\mathrm{LM}}\hat{s}_{\mathrm{LM}},
\end{equation}
where $\hat{s}_{\mathrm{LM}}$ denotes the standardized fluency score.

Beam scores are accumulated across decoding steps:
\begin{equation}
\mathrm{sum}_t = \mathrm{sum}_{t-1} + d_t.
\end{equation}
The base beam score is then defined as
\begin{equation}
\mathrm{score}^{\mathrm{base}}_t =
\begin{cases}
\mathrm{sum}_t / t, & \text{with length normalization},\\
\mathrm{sum}_t, & \text{otherwise}.
\end{cases}
\end{equation}

To encourage diversity, we add repetition-aware penalties:
\begin{equation}
\mathrm{score}_t
=
\mathrm{score}^{\mathrm{base}}_t
+ \lambda_{\mathrm{div}} \mathbf{1}(x_t \in \mathcal{S}_t)
+ \lambda_{\mathrm{ng}} \mathbf{1}(g_t \in \mathcal{G}_t),
\end{equation}
where $\mathcal{S}_t$ is the set of previously selected tokens, $\mathcal{G}_t$ is the set of observed $n$-grams, and $g_t$ denotes the newly formed $n$-gram at step $t$. We retain the top-$W/G$ hypotheses per group, yielding a candidate set $\mathcal{X}$ for the final reconstruction stage.

\subsection{Stage III: Gradient-Space Sparse Reconstruction}
\label{Stage:III}

Given the candidate sentences $\mathcal{X}=\{x^{(1)},\ldots,x^{(M)}\}$ from Stage~II, 
we compute a gradient atom for each candidate:
\[
\bm{g}_i := \bm{\nabla_\theta \mathcal{L}}(f_\theta(x_i), y_i).
\]
While this definition depends on the candidate label $y_i$, the true labels are unavailable at reconstruction time. 
We therefore compute candidate gradients using a fixed surrogate label. 
Empirically, this approximation has negligible impact on candidate ranking in Stage~III (see Appendix~\ref{app:surrogate-label} for details).

To reduce redundancy, we cluster beam outputs by ROUGE-L similarity and retain one representative per cluster.

We then model the observed gradient as a sparse combination over the candidate set:
\begin{equation}
\bm{g}_{\text{mix}}
\approx
\sum_{i\in\mathcal{S}} \alpha_i \bm{g}_i,
\end{equation}
where $\mathcal{S}\subseteq \{1,\ldots,M\}$ is an unknown support set with $|\mathcal{S}|=B$. Intuitively, although Stage~II may generate many plausible candidates, only a small subset should contribute to the true aggregated gradient. We recover this support using Orthogonal Matching Pursuit (OMP).

At iteration $t$, given the current residual $\bm{r}_t$, OMP selects the most aligned candidate:
\begin{equation}
i_t
=
\arg\max_i
\frac{|\langle \bm{g}_i,\bm{r}_t\rangle|}{\|\bm{g}_i\|_2}.
\end{equation}
The active coefficients are then updated by ridge-regularized least squares:
\begin{equation}
\alpha_{\mathcal{S}}
=
\arg\min_{\alpha}
\left\|
\bm{g}_{\text{mix}}-\sum_{j\in\mathcal{S}}\alpha_j \bm{g}_j
\right\|_2^2
+\lambda \|\alpha\|_2^2.
\end{equation}
The residual is updated as
\begin{equation}
\bm{r}_{t+1}
=
\bm{g}_{\text{mix}}-\sum_{j\in\mathcal{S}}\alpha_j \bm{g}_j.
\end{equation}
The procedure terminates when $\|\bm{r}_t\|_2 < \varepsilon$ or $|\mathcal{S}| = B$. The resulting support identifies the candidate sentences whose gradients best explain the observed mixture.The full algorithm is provided in Algorithm~\ref{alg:stage3}.

\subsection{Summary}
\label{sec:theoretical_analysis}

SOMP combines geometric candidate filtering with sparse residual fitting. In Stages~I and~II, head-wise structure is used as a soft constraint to prune implausible hypotheses while preserving a high-recall candidate set. Stage~III then resolves the remaining ambiguity by selecting a sparse subset of candidate gradients that best reconstructs $\bm{g}_{\text{mix}}$. Overall, SOMP is best understood as a structured reconstruction framework rather than a method with worst-case recovery guarantees.

\section{Experimental Evaluation}
\label{sec:ExperimentEval}
We evaluate SOMP across multiple models, datasets, and training protocols along four complementary dimensions. First, we compare against prior text-based inversion methods to assess overall reconstruction quality under aggregated gradients. Second, we vary batch size to test whether SOMP remains effective as sample mixing becomes stronger. Third, we extend the evaluation beyond the standard English FedSGD setting, including a multilingual test on other languages and FedAvg experiments, to examine whether the method relies on English-specific tokenization patterns or on the single-step optimization structure of FedSGD. Finally, we perform ablation studies to identify which stages of the pipeline are most responsible for the gains.
\begin{table*}[t]
\centering
\small
\renewcommand{\arraystretch}{1.1}

\begin{tabular}{c p{0.26\linewidth} p{0.26\linewidth} p{0.26\linewidth}}
\toprule
\textbf{$B$} & \textbf{Reference Inputs} & \textbf{DAGER Reconstruction} & \textbf{SOMP Reconstruction} \\
\midrule




  &
If JognThaw had never ... is absolute must include. &
If JognThaw had never ... is absolute must include. &
If JognThaw had never ... is absolute must include. \\

 &
This sequel is quite ... lack of choreography. &
This sequel is quite ... \hl{movie is} \hl{lack of money}. &
This sequel is quite ... lack of choreography.\hl{.} \\

4 &
I saw this movie when ... in another movie. &
I saw this movie when ... in another movie. &
I saw this movie when ... in another movie. \\

 &
I won't ask you ... give her any money. &
I saw this movie when ... in \hl{another movie}. &
I won't ask you ... give her any money. \\
\midrule

 &
Jill Dunne (played by Mitzi Kapture), is an ... <br/><br/>3 stars. &
Jill Dunne (played by Mitzi Kapture),is an\hl{... nothing}\hl{happens!} &
Jill Dunne (played by Mitzi Kapture), is an ... Jill.\hl{<br/>}3 stars. \\

  &
\multicolumn{1}{c}{$...$} &
\multicolumn{1}{c}{$...$} &
\multicolumn{1}{c}{$...$} \\

8 &
If the term itself were not geographically ... different accents. &
If the term itself were not ... \hl{Ned Kelly geographic Ned Kelly} &
If the term itself were not geographically ... different accents. \\

 &
SWING! It's an important film because ... it's really, really bad! &
SWING! It's an important film because ... \hl{it's real real real}\hl{real} &
SWING! It's an important film because ...it's really\hl{bad!!!} \\

\bottomrule
\end{tabular}

\caption{
Reconstruction examples under aggregated gradients at $B=2,4,8$. Highlighted spans mark incorrect tokens. As $B$ increases, DAGER degrades faster, while SOMP remains more faithful to the reference inputs.
}

\label{tab:qualitative_reconstruction}
\end{table*}

\subsection{Experimental Setup.}
We study the standard honest-but-curious server setting under FedSGD. The attacker observes only the aggregated gradient
\(
\bm{g}_{\mathrm{mix}}
\)
computed over a batch of size \(B\), and does not have access to individual examples or per-sample gradients. Unless otherwise stated, gradients are extracted from the first-layer query projections.

\paragraph{Models and datasets.}
We evaluate \textbf{SOMP} on three autoregressive LLMs: \texttt{GPT-2}~\cite{Radford2019LanguageMA}, \texttt{GPT-J-6B}~\cite{gpt-j}, and \texttt{Qwen3-8B}~\cite{yang2025qwen3technicalreport}. Reconstruction experiments are conducted on \texttt{CoLA}~\cite{warstadt-etal-2019-neural}, \texttt{SST-2}~\cite{socher-etal-2013-recursive}, and \texttt{stanfordNLP/imdb}~\cite{maas-etal-2011-learning}. Inputs are truncated to 512 tokens. 

\paragraph{Baselines and metrics.}
We compare against representative prior text-based inversion methods: DAGER, LAMP, and GRAB. All baselines are reimplemented under the same aggregation protocol and evaluation pipeline. Reconstruction quality is measured using sequence-level \textbf{ROUGE-1}, \textbf{ROUGE-2}, and \textbf{ROUGE-L}~\cite{lin-2004-rouge}, averaged over 50 random batches and reported as mean \(\pm\) standard deviation. For readability, the main text reports \textbf{ROUGE-L}; full ROUGE-1/2/L results are provided in Appendix~\ref{app:MainExperiment} and follow the same overall setting.

Our main experiments focus on standard aggregated-gradient training rather than formal DP mechanisms such as DP-SGD. We include an exploratory Gaussian-noise study in Appendix~\ref{app:dp}, while a full DP-SGD evaluation with clipping and privacy accounting is left for future work.

\begin{figure*}[t]
    \centering
    \includegraphics[width=0.9\textwidth]{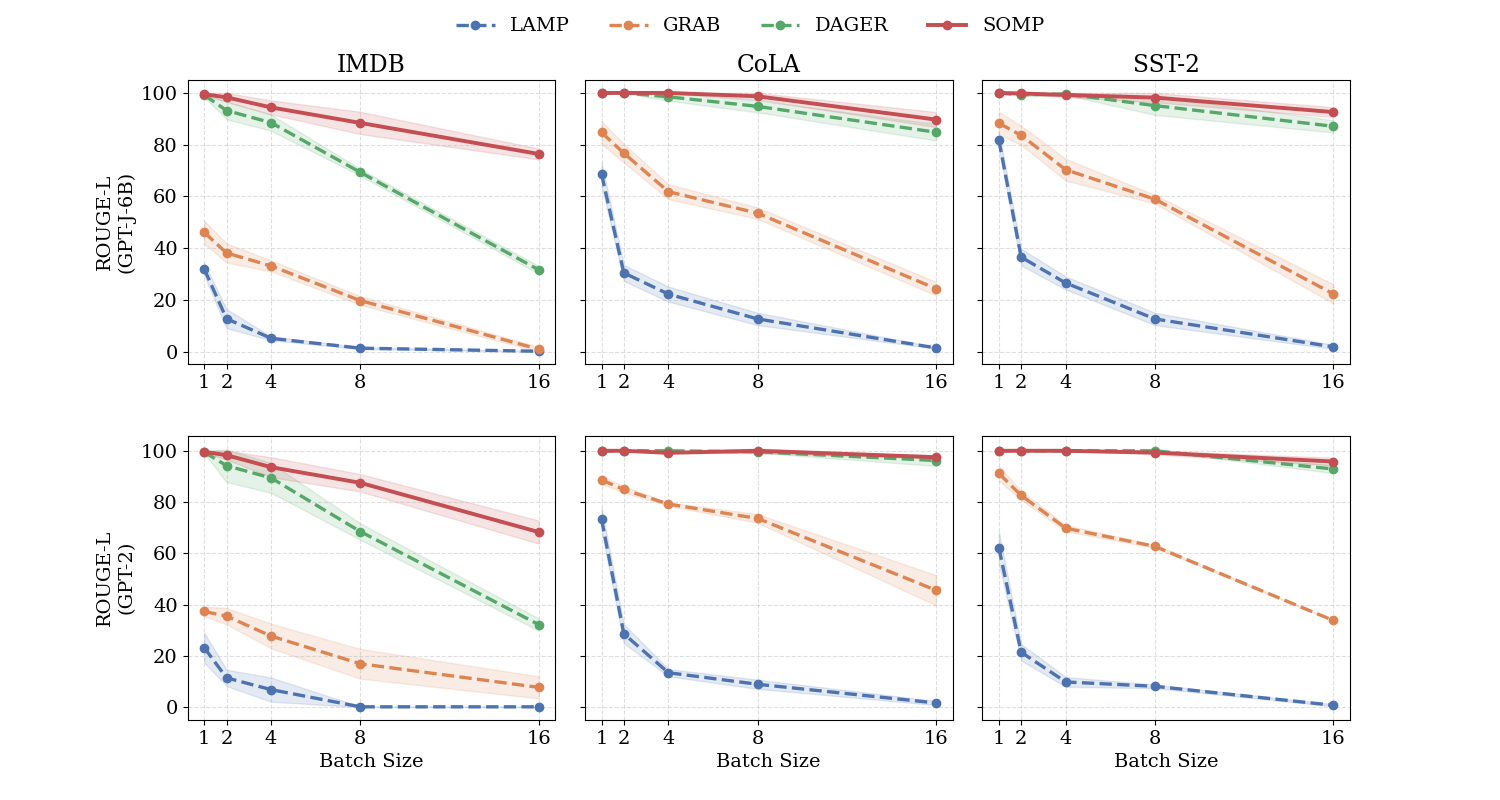}
    \caption{
    Reconstruction performance (ROUGE-L) under aggregated gradients at batch sizes B = 1, 2, 4, 8, and 16 on IMDB, CoLA, and SST-2 using GPT-J-6B (top) and GPT-2 (bottom). SOMP shows the largest gains on IMDB, whose longer inputs make multi-sample inversion harder than on shorter datasets such as CoLA and SST-2.
    }
    \label{fig:mainRes_fig}
\end{figure*}

\subsection{Reconstruction Performance and Comparisons}
\label{sec:main_results}
Figure~\ref{fig:mainRes_fig} and Table~\ref{tab:rougel_main} summarize the main comparison against prior text-based gradient inversion methods. The overall pattern is clear: SOMP achieves the strongest reconstruction quality once gradients are mixed across multiple samples, and its advantage becomes more pronounced as batch size increases. The largest gains appear on \texttt{IMDB}, where long inputs make cross-sample interference especially severe. Table~\ref{tab:qualitative_reconstruction} provides representative qualitative examples comparing DAGER and SOMP. Under mixed gradients, DAGER often merges samples with similar prefixes, repeats locally plausible tokens, or drifts toward nearby embedding-space neighbors. The resulting outputs can remain superficially fluent while failing to separate the underlying samples correctly. In contrast, SOMP more reliably disentangles the mixed signals and reconstructs the original inputs. Additional qualitative examples comparing all methods are provided in Appendix~\ref{ReconstructionExample}.

\noindent\textbf{The main difficulty comes from increasing batch size, not from sequence length alone.}
\texttt{IMDB} is the most challenging benchmark in our study because it combines long inputs with cross-sample mixing. As shown in Table~\ref{tab:rougel_main}, reconstruction quality generally degrades as \(B\) increases, but prior approaches degrade much more sharply. For example, on GPT-J-6B, DAGER drops from \(58.4\) at \(B=4\) to \(39.2\) at \(B=8\), whereas SOMP still reaches \(73.5\) at \(B=8\). This gap widens further at \(B=16\).

Importantly, the full results in Appendix~\ref{app:MainExperiment} show that DAGER remains highly effective at \(B=1\) on \texttt{IMDB}. This indicates that long sequences alone are not the main bottleneck for search-based inversion. The real challenge emerges when multiple long documents are mixed into a single observed gradient, which is exactly the regime where SOMP provides its largest gains.  On shorter-sequence benchmarks such as \texttt{CoLA} and \texttt{SST-2}, inversion remains comparatively easier, and both DAGER and SOMP perform strongly at small and moderate batch sizes. Even in these easier settings, however, SOMP degrades more gracefully as \(B\) grows. This shows that SOMP is not mainly designed to maximize gains in easy cases, but to remain effective where prior attacks become brittle.

\begin{figure}[t]
    \centering
    \includegraphics[width=0.9\linewidth]{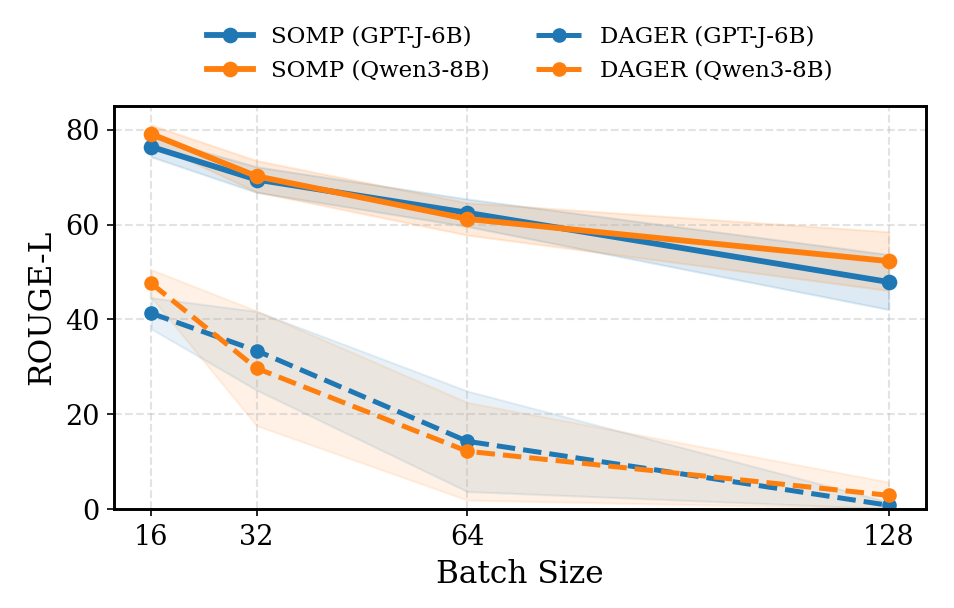}
    \caption{Reconstruction performance (ROUGE-L) on GPT-J-6B and Qwen3-8B over larger batch sizes on \texttt{IMDB}. SOMP remains effective substantially beyond the regime where prior methods degrade sharply.}
    \label{fig:scaling}
\end{figure}

\noindent\textbf{SOMP remains effective at substantially larger batch sizes.}
Figure~\ref{fig:scaling} extends the comparison to much larger batch sizes on \texttt{IMDB}. We can see that DAGER degrades rapidly as \(B\) increases. In contrast, SOMP continues to recover meaningful text even at batch sizes far beyond the range where earlier methods begin to collapse. The same trend appears for both \texttt{GPT-J-6B} and \texttt{Qwen3-8B}, suggesting that the advantage of SOMP is not specific to a single architecture family.

\begin{table}[t]
\centering
\small
\setlength{\tabcolsep}{3pt}
\renewcommand{\arraystretch}{1.15}
\caption{Reconstruction performance (ROUGE-L) under aggregated gradients at $B=4,8,16$ on GPT-J-6B.}
\label{tab:rougel_main}
\begin{tabular}{l l c c c}
\hline
\textbf{Dataset} & \textbf{Method} & $B=4$ & $B=8$ & $B=16$ \\
\hline
 & LAMP & $1.2\pm0.8$ & $0.3\pm0.3$ & $0.1\pm0.1$ \\
 & GRAB & $8.6\pm2.4$ & $2.0\pm0.6$ & $0.9\pm0.9$ \\
IMDB
 & DAGER & $58.4\pm2.4$ & $39.2\pm0.8$ & $31.6\pm1.5$ \\
 & \textbf{SOMP}  & \textbf{89.3}$\pm$\textbf{1.9} & \textbf{73.5}$\pm$\textbf{5.1} & \textbf{76.4}$\pm$\textbf{2.1} \\
\hline
 & LAMP & $10.6\pm1.8$ & $6.3\pm0.7$ & $1.4\pm0.3$ \\
 & GRAB & $42.6\pm2.4$ & $35.1\pm2.1$ & $24.3\pm2.6$ \\
CoLA
 & DAGER & $92.9\pm3.1$ & $87.8\pm4.5$ & $84.8\pm3.2$ \\
 & \textbf{SOMP}  & \textbf{96.3}$\pm$\textbf{3.7} & \textbf{94.5}$\pm$\textbf{3.3} & \textbf{89.7}$\pm$\textbf{2.9} \\
\hline
 & LAMP & $12.9\pm1.2$ & $7.2\pm0.7$ & $1.8\pm0.8$ \\
 & GRAB & $43.3\pm2.3$ & $36.4\pm2.5$ & $22.3\pm3.7$ \\
SST-2
 & DAGER & $100.0\pm0.0$ & $92.9\pm1.6$ & $87.1\pm2.4$ \\
 & \textbf{SOMP}  & \textbf{100.0}$\pm$\textbf{0.0} & \textbf{95.8}$\pm$\textbf{1.5} & \textbf{92.6}$\pm$\textbf{1.9} \\
\hline
\end{tabular}
\end{table}

\begin{table}[t]
\centering
\small
\setlength{\tabcolsep}{3pt}
\renewcommand{\arraystretch}{1.05}
\caption{Average per-batch runtime in seconds. Bw denotes beam width.}
\label{tab:runtime_gpu}
\begin{tabular}{l l rrrrr}
\hline
\textbf{Dataset} & \textbf{Method} &
\textbf{B=1} & \textbf{B=2} & \textbf{B=4} & \textbf{B=8} &
\textbf{B=16} \\
\hline
\multicolumn{7}{c}{\textbf{GPT-2 (runtime measured on RTX 5060)}} \\
\hline
     & DAGER        &  3 &  6 & 10  & 19  & 36  \\
CoLA & SOMP(Bw=4)   &  7 & 10 & 13  & 21  & 26  \\
     & SOMP(Bw=8)   &  7 & 10 & 14  & 24  & 31  \\
     & SOMP(Bw=16)  &  8 & 13 & 21  & 33  & 41  \\
\hline
      & DAGER       &  3 &  4 &  9  & 18  & 31  \\
SST-2 & SOMP(Bw=4)  &  6 &  9 & 13  & 22  & 30  \\
      & SOMP(Bw=8)  &  7 & 11 & 14  & 26  & 37  \\
      & SOMP(Bw=16) &  8 & 12 & 30  & 40  & 52  \\
\hline
     & DAGER        &  61 & 118 & 248 & 467 & 1127 \\
IMDB & SOMP(Bw=4)   &  48 &  96 & 191 & 359 &  672 \\
     & SOMP(Bw=8)   &  58 & 117 & 231 & 448 &  968 \\
     & SOMP(Bw=16)  &  75 & 143 & 289 & 572 & 1308 \\
\hline
\end{tabular}
\end{table}
\begin{figure*}[h]
    \centering
    \includegraphics[width=0.9\linewidth]{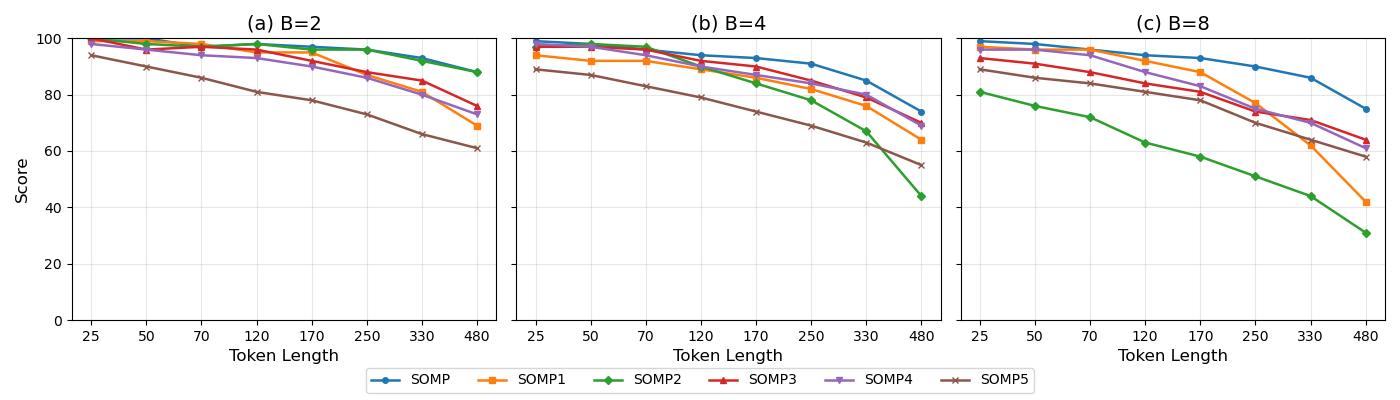}
    \caption{Ablation at batch sizes ($B=2,4,8$). Variants remove LM prior (SOMP1), OMP (SOMP2), sparsity scoring (SOMP3), cross-head consistency (SOMP4), and subspace-distance scoring (SOMP5).}
    \label{fig:ablation}
\end{figure*}


\subsection{Efficiency and Scalability}

SOMP improves efficiency by filtering and scoring candidates in informative head-wise subspaces, rather than repeatedly verifying them in the full dimensional space. This matters most when the candidate set grows large, since full-space verification becomes the main cost for search-based baselines such as DAGER.
Table~\ref{tab:runtime_gpu} reports measured per-batch runtime under several beam widths. On short-sequence benchmarks such as \texttt{CoLA} and \texttt{SST-2}, SOMP may incur some overhead from Stage~III sparse reconstruction. On \texttt{IMDB}, however, where sequences are longer and the search space is much larger, SOMP shows a clearly more favorable runtime profile.
We note that SOMP is not designed to be uniformly faster in every setting. In easier regimes, the added OMP stage can make it comparable to or slightly slower than simpler baselines. In long-sequence, larger-batch settings, however, SOMP becomes increasingly competitive and is often faster, while also achieving substantially better reconstruction quality.

\subsection{Multilingual Evaluation}
\label{multilingualEval}
We next test whether SOMP generalizes beyond English. Across five languages, SOMP maintains strong reconstruction quality without language-specific tuning, suggesting that its effectiveness does not depend on English-specific tokenization patterns. As shown in Table~\ref{tab:multilingual-bs} of Appendix~\ref{MultilingualExperiment}, performance remains strongest on English, German, and Chinese, while French and Italian degrade more noticeably as batch size increases. A likely reason is heavier subword fragmentation, which leads to longer tokenized sequences and a larger decoding search space. Even so, SOMP remains effective across all evaluated languages, indicating that it exploits structural properties of aggregated transformer gradients that persist across languages.

\subsection{Reconstruction under FedAvg}
\label{sec:fedavg_main}
We also evaluate SOMP under \textbf{FedAvg}, where clients perform multiple local updates before aggregation. This setting is more realistic than one-step FedSGD and more challenging for inversion, as the attacker observes only the final model update after local training (see Appendix~\ref{app:fedavg}). The results show that FedAvg does not eliminate leakage: across local epochs, learning rates, and mini-batch sizes, SOMP still recovers substantial information from the observed update. Overall, these experiments indicate that the leakage exploited by SOMP is not specific to single-step FedSGD, but can persist under practical federated optimization.

\subsection{Ablation Study}
We ablate the main components of SOMP in Figure~\ref{fig:ablation}, including the LM prior, OMP-based sparse reconstruction, sparsity scoring, cross-head consistency, and subspace-distance scoring.
Removing OMP causes the largest drop, especially at larger batch sizes and longer sequence lengths, showing that sparse reconstruction is the key step for resolving cross-sample ambiguity in the highly mixed regime. The other components also contribute consistently: the LM prior and cross-head consistency matter more as batch size increases, while sparsity scoring and subspace-based filtering help suppress noisy candidates earlier in the pipeline. Overall, the gains of SOMP come from the interaction of its stages rather than from any single heuristic.

\section{Conclusion}

We presented SOMP, a subspace-guided and sparsity-driven framework for
reconstructing textual inputs from aggregated gradients of large language
models.
By exploiting the multi-head structure of transformer gradients, SOMP combines
geometric filtering, language-model--guided decoding, and sparse gradient-space
reconstruction. Experimental results show that SOMP enables accurate and stable multi-sample
reconstruction under aggregated-gradient settings, particularly for long
sequences and large batch sizes.
Overall, SOMP provides a principled and scalable approach to gradient inversion
in practical training scenarios.
\newpage
\section{Limitations}
SOMP relies on structural regularities in first-layer query gradients and uses
head-wise decomposition to reduce the search space. Although modern
transformers may adopt alternative attention designs such as grouped-query
attention (GQA), our experiments on Qwen3-8B show that performance remains comparable under this architecture. A more challenging setting would
be architectures where head-level gradient patterns are substantially less
separable.

In addition, Stage~I requires maintaining a high-recall token pool. Under
severely distorted, compressed, or noisy gradients, relevant tokens may be
pruned too early during candidate filtering, affecting downstream
reconstruction. From an efficiency perspective, while head-wise factorization reduces the cost
of candidate verification, the OMP-based sparse reconstruction in Stage~III
introduces additional overhead. As a result, SOMP may be slower on short
sequences and small batch sizes, but becomes increasingly competitive as
sequence length and aggregation level increase.

Finally, our main evaluation does not study formal differential privacy
mechanisms such as DP-SGD. While we include initial experiments with
additive Gaussian noise in the Appendix~\ref{app:dp}, these results do not constitute a
systematic evaluation of differential privacy, since they do not incorporate
the full DP-SGD pipeline with gradient clipping and privacy accounting.
A more complete study under noisy, clipped, or DP-SGD-perturbed gradients
remains future work.

\section{Ethical Considerations}

This work investigates the feasibility of reconstructing textual inputs from aggregated gradients, highlighting potential privacy risks in federated and distributed training of large language models. Our goal is to better understand the conditions under which gradient leakage may occur, in order to inform the design of safer and more privacy-aware training protocols.

We do not advocate the misuse of gradient inversion techniques. Instead, our findings emphasize the importance of established privacy safeguards, such as secure aggregation and differential privacy.

\bibliography{custom}

\appendix

\appendix

\section{Head-wise Sparsity Decomposition}
\label{app:head_sparsity}

In this appendix, we provide a formal justification showing that measuring sparsity globally or in a head-wise manner is equivalent under a block-wise decomposition of feed-forward network (FFN) parameters.

\subsection{Additivity of Head-wise Sparsity}

\paragraph{Proposition A.1 (Additivity of Head-wise Sparsity).}
Let $G \in \mathbb{R}^{d \times D}$ be a gradient-related matrix associated with a feed-forward network layer. Assume that $G$ is partitioned column-wise into $H$ disjoint blocks:
\begin{equation}
G = [\, G^{(1)}, G^{(2)}, \dots, G^{(H)} \,],
\end{equation}
where $G^{(h)} \in \mathbb{R}^{d \times D_h}$ and $\sum_{h=1}^{H} D_h = D$.

For a token embedding $e \in \mathbb{R}^{d}$, define its projected response
\begin{equation}
g = G^\top e \in \mathbb{R}^{D},
\end{equation}
and correspondingly the head-wise responses
\begin{equation}
g^{(h)} = (G^{(h)})^\top e \in \mathbb{R}^{D_h}.
\end{equation}

Then the global sparsity of $g$, measured by the number of non-zero entries, satisfies
\begin{equation}
\| g \|_0 = \sum_{h=1}^{H} \| g^{(h)} \|_0 .
\end{equation}

An analogous decomposition holds for threshold-based sparsity measures where entries with magnitude below a fixed threshold are treated as zero.

\paragraph{Proof.}
By construction, the blocks $G^{(h)}$ correspond to disjoint column subsets of $G$. Consequently, the vectors $g^{(h)}$ occupy disjoint coordinate subsets of $g$. Since sparsity measures such as the $\ell_0$ count (or thresholded variants) operate independently on each coordinate, the total number of non-zero entries in $g$ is exactly the sum of the non-zero entries across all head-wise components:
\begin{equation}
\| g \|_0 = \sum_{h=1}^{H} \| g^{(h)} \|_0 .
\end{equation}

\subsection{Discussion}

Proposition~A.1 shows that evaluating sparsity in a head-wise manner is mathematically equivalent to evaluating sparsity on the full projected response under a block-wise decomposition of FFN parameters. Therefore, adopting a head-aware sparsity formulation does not introduce additional assumptions, nor does it alter the underlying sparsity-based filtering objective. Instead, it provides a structured reformulation aligned with the architectural organization of modern transformer FFNs and facilitates more interpretable and numerically stable sparsity estimation.

\section{Effect of Surrogate Labels in Stage III}
\label{app:surrogate-label}
The Stage III gradient atom is defined as
\[
g_i = \nabla_\theta \mathcal{L}(f_\theta(x_i), y_i).
\]
Since the true labels of candidate sentences are unknown at reconstruction time, we compute candidate gradients using a fixed surrogate label.

\paragraph{Gradient decomposition.}
For a softmax classifier with cross-entropy loss,
\[
\nabla_\theta \mathcal{L}(x,y)
=
J_\theta(x)^\top (p-y),
\]
where \(J_\theta(x)=\frac{\partial z}{\partial\theta}\) is the Jacobian of the logits and \(p=\mathrm{softmax}(z)\).
Thus the gradient consists of an input-dependent Jacobian \(J_\theta(x)\) and a label-dependent coefficient vector \((p-y)\).

If the true label \(y\) is replaced by a surrogate label \(\tilde y\), the gradient becomes
\[
\tilde g(x) = J_\theta(x)^\top (p-\tilde y).
\]
Both \(g(x,y)\) and \(\tilde g(x)\) therefore lie in the same subspace
\[
\mathrm{colspan}(J_\theta(x)^\top),
\]
meaning that replacing the label preserves the input-dependent gradient structure and only changes the coefficient vector.

\paragraph{Implication for Stage III.}
Stage III ranks candidates using gradient alignment scores
\[
\langle g_, r_t\rangle .
\]
Replacing the true label only modifies the coefficient vector in \(g_r\) while leaving the Jacobian interaction structure unchanged. 
Consequently, the dominant input-dependent gradient structure is preserved, and in practice the resulting candidate ranking is only weakly affected by the choice of label.

\section{Hyperparameters}
\label{app:hyperparameters}

This appendix summarizes the hyperparameters used in \textbf{SOMP}. We distinguish between batch-size-independent hyperparameters, which are fixed across all experiments, and batch-size-dependent hyperparameters, which vary with the aggregation level $B$.

The batch-size-dependent parameters reflect the increased ambiguity and sparsity requirements induced by larger gradient mixtures. Importantly, these parameters follow a fixed strategy shared across datasets, languages, and models, rather than being tuned per task or per instance.


\begin{table}[h]
\centering
\small
\setlength{\tabcolsep}{4pt}
\renewcommand{\arraystretch}{1.1}
\begin{tabular}{lll}
\hline
\textbf{Category} & \textbf{Hyperparameter} & \textbf{Value} \\
\hline
Stage I &
$\lambda_1$ & $0.8$ \\
&
$\lambda_2$ & $0.5$ \\
&
$\lambda_3$ & $0.5$ \\
&
$\tau^{(h)}_m$ & $0.5\times$ head-wise median \\
\hline
Stage II 
&
$\beta_{\text{LM}}$ & $0.33$ \\
&
$\lambda_{\text{div}}$ & $0.15$ \\
&
$\lambda_{\text{ng}}$ & $0.2$ \\
\hline
Stage III &
$\lambda$ & $1e{-}3$ \\
&
$\epsilon$ & $1e{-}4 \cdot \|\bm{g}_{\text{mix}}\|_2$ \\
&
$\tau_{\text{cluster}}$ & $0.8 \times$ Rouge-L \\
\hline
Model / Exp. &
$H$ & model-specific \\
&
$P$ & 512 \\
\hline
\end{tabular}
\caption{Batch-size-independent hyperparameters used in SOMP.}
\end{table}


\begin{table}[h]
\centering
\small
\setlength{\tabcolsep}{4pt}
\renewcommand{\arraystretch}{1.1}
\begin{tabular}{lllll}
\hline
\textbf{Hyperparameter} &
\textbf{$B{=}1$} &
\textbf{$B{=}4$} &
\textbf{$B{=}8$} &
\textbf{$B{=}16{+}$} \\
\hline
Token pool size $k$ & 960 & 1600 & 2400 & 3200 \\
Informative heads $|H_{\text{act}}|$ & $H/4$ & $H/4$ & $H/3$ & $H/3$ \\
Sparsity heads $|H_{\text{top}}|$ & $H/6$ & $H/6$ & $H/4$ & $H/4$ \\
Beam width $W$ & 2 & 4 & 6 & 12 \\
Number of beam groups $G$ & 1 & 4 & 8 & 16 \\
OMP iterations (max) & $B$ & $B$ & $B$ & $B$ \\
\hline
\end{tabular}
\caption{Batch-size-dependent hyperparameters as a function of aggregation level $B$.}
\end{table}

\paragraph{Discussion.}
As the batch size increases, the aggregated gradient represents a mixture of a larger number of samples, increasing both combinatorial ambiguity and sparsity requirements. Accordingly, SOMP increases the size of the candidate token pool and the breadth of sentence-level search while maintaining fixed geometric and sparsity weighting. This strategy enables stable reconstruction across a wide range of aggregation levels without per-task hyperparameter tuning.

\begin{algorithm}[tbp]
\caption{Stage I: Head-Structured Token Pooling}
\label{alg:stage1}
\begin{algorithmic}[1]
\Require Aggregated gradient $g_{\text{mix}}$, vocabulary $V$
\Ensure Filtered token pool $\mathcal{P}$
\Function{TokenPooling}{$g_{\text{mix}}, V$}
    \State Decompose $g_{\text{mix}}$ into $\{G^{(h)}\}_{h=1}^{H}$
    \State Compute $\mathcal{R}_Q^{(h)} \gets \mathrm{Span}(G^{(h)})$
    \State $\mathcal{P} \gets \emptyset$
    \For{$v \in V$}
        \State Compute $s_{\text{sub}}(v)$, $s_{\text{cons}}(v)$, $s_{\text{sparse}}(v)$
        \State $s_{\text{total}} \gets \lambda_1 s_{\text{sub}} + \lambda_2 s_{\text{cons}} - \lambda_3 s_{\text{sparse}}$
        \If{$v \in \mathrm{Top}\text{-}k$}
            \State $\mathcal{P} \gets \mathcal{P} \cup \{v\}$
        \EndIf
    \EndFor
    \State \Return $\mathcal{P}$
\EndFunction
\end{algorithmic}
\end{algorithm}

\begin{algorithm}[tbp]
\caption{Stage II: Diverse Beam Decoding}
\label{alg:stage2}
\footnotesize
\begin{algorithmic}[1]
\Require Global pool $\mathcal{P}$, subspaces $\mathcal{R}$, groups $G$
\Ensure Candidate set $\mathcal{X}$
\Function{DiverseBeamSearch}{$\mathcal{P}, \mathcal{R}$}
    \State Detect candidate lengths $\mathcal{L} \gets \{L_1, \dots \}$
    \State $\mathcal{X}_{raw} \gets \emptyset$
    \For{$L \in \mathcal{L}$}
        \For{$t \gets 1 To L$}
            \State $\mathcal{P}_t \gets \{ v \in \mathcal{P} \mid s_{\text{sub}}(v,t) < \tau_{\text{pos}} \}$
        \EndFor
        \State Initialize beam groups $\mathcal{B}_1, \dots, \mathcal{B}_G$
        \For{$t \gets 1 To L$}
            \For{$g \in 1 \dots G$}
                \State Expand $\mathcal{B}_g$ using $\mathcal{P}_t$
                \State $S \gets \text{score}^{\text{base}}_t + \lambda_{\text{div}} \mathbf{1}(x_t \in S_t) + \lambda_{\text{ng}} \mathbf{1}(g_t \in G_t)$
                \State Prune to top-$W/G$
            \EndFor
        \EndFor
        \State $\mathcal{X}_{raw} \gets \mathcal{X}_{raw} \cup \bigcup \mathcal{B}_g$
    \EndFor
    \State $\mathcal{C} \gets \text{Cluster}(\mathcal{X}_{raw})$
    \State \Return $\mathcal{X} \gets \{ \text{Rep}(c) \mid c \in \mathcal{C} \}$
\EndFunction
\end{algorithmic}
\end{algorithm}

\subsection{Hyperparameter Sensitivity Analysis}
\label{app:sensitivityAnalysis}

We analyze the sensitivity of SOMP to several key hyperparameters used in the three-stage reconstruction pipeline. In particular, we examine the influence of parameters controlling token pooling, beam search, and sparse reconstruction.

Unless otherwise stated, experiments are conducted on the IMDB dataset using GPT-J with batch size $B=8$. For each experiment, we vary one hyperparameter while keeping all others fixed according to the hyperparameter settings above. Reconstruction quality is measured using ROUGE-L.

Overall, SOMP behaves as expected under hyperparameter variation, but remains stable under moderate changes. Small to moderate deviations in parameter values do not lead to abrupt degradation in reconstruction performance, indicating that the method does not rely on fragile tuning.

\begin{table}[t]
\centering
\small
\setlength{\tabcolsep}{4pt}
\renewcommand{\arraystretch}{1.1}
\caption{Hyperparameter sensitivity analysis of SOMP. Each parameter is varied while keeping the others fixed.}
\label{tab:sensitivity}
\begin{tabular}{l c c}
\hline
Parameter & Value & ROUGE-L \\
\hline

\multicolumn{3}{c}{\textit{Stage I: Token Pooling}} \\
\hline
Token pool size $k$ & 800  & 59.7 \\
                   & 1200 & 71.5 \\
                   & 1600 & 73.4 \\
                   & 2000 & 74.5 \\
                   & 2400 & 73.3 \\

$\lambda_1$ & 0.6 & 74.7 \\
            & 0.8 & 67.3 \\
            & 1.0 & 53.4 \\

$\lambda_2$ & 0.3 & 68.4 \\
            & 0.5 & 74.5 \\
            & 0.7 & 61.7 \\

\hline
\multicolumn{3}{c}{\textit{Stage II: Beam Decoding}} \\
\hline

Beam width $W$  &    2  & 37.1 \\
                &    4  & 59.9 \\
                &    6  & 66.4 \\
                &    8  & 72.7 \\
                &    12 & 79.3 \\

$\lambda_{\text{div}}$ & 0.1 & 62.2 \\
                      & 0.15 & 74.7 \\
                      & 0.2 & 69.5 \\

$\lambda_{\text{ng}}$ & 0.1 & 64.8 \\
                     & 0.2 & 74.8 \\
                     & 0.3 & 71.5 \\

\hline
\multicolumn{3}{c}{\textit{Stage III: Sparse Reconstruction}} \\
\hline

$\lambda$ & $5\times10^{-4}$ & 69.8 \\
          & $1\times10^{-3}$ & 74.2 \\
          & $5\times10^{-3}$ & 71.6 \\
          & $1\times10^{-2}$ & 73.7 \\

$\epsilon$ & $5\times10^{-5}$ & 76.5 \\
           & $1\times10^{-4}$ & 73.9 \\
           & $5\times10^{-4}$ & 68.4 \\

$\tau_{cluster}$ & 0.6 & 67.0 \\
                 & 0.8 & 73.5 \\
                 & 1.0 & 81.4 \\

\hline
\end{tabular}
\end{table}

\paragraph{Discussion.}
The results show that SOMP is influenced by hyperparameter choices, as larger token pools or beam widths generally improve candidate coverage at the cost of increased computation. However, performance varies smoothly across a reasonable range of parameter values. In particular, moderate changes in $k$, $W$, or $\lambda$ do not cause catastrophic degradation in reconstruction quality. These observations suggest that SOMP does not depend on fragile hyperparameter tuning and remains robust under typical parameter variations.

\section{Algorithmic Details}
\label{app:algorithms}

This appendix provides pseudocode for the three stages of SOMP, complementing the high-level descriptions in Section~\ref{sec:methodology}. Algorithm~\ref{alg:stage1} details head-structured token pooling, Algorithm~\ref{alg:stage2} presents geometry-driven diverse beam decoding, and Algorithm~\ref{alg:stage3} describes gradient-space sparse reconstruction via Orthogonal Matching Pursuit (OMP).

The pseudocode is intended to improve clarity and reproducibility and does not introduce additional assumptions beyond those already discussed in the main text.

\begin{algorithm}[tbp]
\caption{Stage III: OMP Reconstruction}
\label{alg:stage3}
\footnotesize
\begin{algorithmic}[1]
\Require $g_{\mathrm{mix}}$, candidates $\mathcal{X}$, batch size $B$
\Ensure Reconstructed batch $\mathcal{S}^*$
\Function{OMP}{$g_{\mathrm{mix}}, \mathcal{X}$}
    \State Dictionary $\mathcal{D} \gets \{ \nabla \mathcal{L}(x) \mid x \in \mathcal{X} \}$
    \State $r_0 \gets g_{\mathrm{mix}}, \quad \Lambda_0 \gets \emptyset$
    \For{$k = 1 \to B$}
        \State $j^* \gets \arg\max_{j} \frac{|\langle g_j, r_{k-1} \rangle|}{\|g_j\|_2}$
        \State $\Lambda_k \gets \Lambda_{k-1} \cup \{j^*\}$
        \State $\alpha^* \gets 
            \arg\min_{\alpha}\,
            \bigl\| g_{\mathrm{mix}} - \sum_{i \in \Lambda_k} \alpha_i g_i \bigr\|_2^2
            + \lambda \|\alpha\|_2^2$
        \State $r_k \gets 
            g_{\mathrm{mix}} - \sum_{i \in \Lambda_k} \alpha_i^* g_i$
        \If{$\|r_k\| < \epsilon$}
            \State \textbf{break}
        \EndIf
    \EndFor
    \State \Return $\mathcal{S}^* \gets \{\, x_j \mid j \in \Lambda_k \,\}$
\EndFunction
\end{algorithmic}
\end{algorithm}


\begin{table*}[t]
\centering
\small
\setlength{\tabcolsep}{4pt}
\renewcommand{\arraystretch}{1.05}

\begin{tabular}{l l cc cc cc cc}
\hline
\textbf{Dataset} & \textbf{Method} &
\textbf{R-1} & \textbf{R-2} &
\textbf{R-1} & \textbf{R-2} &
\textbf{R-1} & \textbf{R-2} &
\textbf{R-1} & \textbf{R-2} \\
& & \multicolumn{2}{c}{$B=1$} &
\multicolumn{2}{c}{$B=2$} &
\multicolumn{2}{c}{$B=4$} &
\multicolumn{2}{c}{$B=8$} \\
\hline
\multicolumn{10}{c}{\textbf{GPT-J}} \\
\hline
 & LAMP & 31.9$\pm$2.4 & 16.5$\pm$3.1 & 12.7$\pm$3.7 & 7.6$\pm$2.0 & 5.1$\pm$0.7 & 1.4$\pm$0.3 & 1.3$\pm$0.4 & 0.3$\pm$0.3 \\
IMDB & GRAB & 46.1$\pm$4.4 & 33.4$\pm$1.4 & 38.1$\pm$3.7 & 21.7$\pm$0.9 & 33.1$\pm$2.1 & 9.2$\pm$1.3 & 19.7$\pm$1.7 & 2.0$\pm$0.6 \\
 & DAGER & 99.2$\pm$0.8 & 94.3$\pm$3.2 & 93.2$\pm$3.4 & 72.0$\pm$2.8 & 88.5$\pm$3.1 & 58.4$\pm$2.4 & 69.3$\pm$1.3 & 39.2$\pm$0.8 \\
 & \textbf{SOMP} & \textbf{99.5$\pm$0.5} & \textbf{96.5$\pm$2.1} & \textbf{98.3$\pm$1.7} & \textbf{93.6$\pm$3.8} & \textbf{94.4$\pm$2.7} & \textbf{89.3$\pm$1.9} & \textbf{88.4$\pm$4.3} & \textbf{73.5$\pm$5.1} \\
\hline
 & LAMP & 68.6$\pm$4.5 & 42.6$\pm$4.7 & 30.4$\pm$3.1 & 17.8$\pm$1.6 & 22.2$\pm$2.9 & 10.4$\pm$1.5 & 12.6$\pm$2.4 & 6.3$\pm$0.7 \\
CoLA & GRAB & 84.7$\pm$4.4 & 62.5$\pm$2.7 & 76.7$\pm$3.6 & 57.2$\pm$3.8 & 61.8$\pm$2.9 & 42.6$\pm$2.4 & 53.5$\pm$2.2 & 35.1$\pm$2.1 \\
 & DAGER & 100.0$\pm$0.0 & 100.0$\pm$0.0 & 100.0$\pm$0.0 & 100.0$\pm$0.0 & 98.5$\pm$1.5 & 92.9$\pm$3.1 & 94.8$\pm$2.3 & 87.8$\pm$4.5 \\
 & \textbf{SOMP} & \textbf{100.0$\pm$0.0} & \textbf{100.0$\pm$0.0} & \textbf{100.0$\pm$0.0} & \textbf{100.0$\pm$0.0} & \textbf{100.0$\pm$0.0} & \textbf{96.3$\pm$3.7} & \textbf{98.7$\pm$1.3} & \textbf{94.5$\pm$3.3} \\
\hline
 & LAMP & 81.8$\pm$3.0 & 57.8$\pm$6.9 & 36.5$\pm$3.2 & 23.6$\pm$3.8 & 26.5$\pm$2.5 & 12.9$\pm$1.2 & 12.6$\pm$2.4 & 7.2$\pm$0.7 \\
SST-2 & GRAB & 88.3$\pm$4.4 & 66.3$\pm$1.2 & 83.6$\pm$3.7 & 56.5$\pm$4.5 & 70.3$\pm$4.2 & 43.3$\pm$2.3 & 58.9$\pm$1.7 & 36.4$\pm$2.5 \\
 & DAGER & 100.0$\pm$0.0 & 100.0$\pm$0.0 & 99.3$\pm$0.7 & 92.0$\pm$1.8 & 99.6$\pm$0.4 & 90.4$\pm$3.3 & 95.1$\pm$3.6 & 86.8$\pm$2.9 \\
 & \textbf{SOMP} & \textbf{100.0$\pm$0.0} & \textbf{100.0$\pm$0.0} & \textbf{100.0$\pm$0.0} & \textbf{99.8$\pm$0.2} & \textbf{99.2$\pm$0.8} & \textbf{97.6$\pm$2.4} & \textbf{98.2$\pm$1.8} & \textbf{95.9$\pm$2.5} \\
\hline
\end{tabular}

\qquad

\begin{tabular}{l l cc cc cc cc}
\hline
\textbf{Dataset} & \textbf{Method} &
\textbf{R-1} & \textbf{R-2} &
\textbf{R-1} & \textbf{R-2} &
\textbf{R-1} & \textbf{R-2} &
\textbf{R-1} & \textbf{R-2} \\
& & \multicolumn{2}{c}{$B=1$} &
\multicolumn{2}{c}{$B=2$} &
\multicolumn{2}{c}{$B=4$} &
\multicolumn{2}{c}{$B=8$} \\
\hline
\multicolumn{10}{c}{\textbf{GPT-2}} \\
\hline
 & LAMP  & 23.2$\pm$5.8 & 18.8$\pm$8.4 & 11.4$\pm$3.2 & 7.4$\pm$1.6 & 6.8$\pm$4.7 & 2.7$\pm$2.4 & 0.1$\pm$0.1 & 0.1$\pm$0.1 \\
IMDB & GRAB  & 37.3$\pm$1.8 & 29.1$\pm$1.0 & 35.5$\pm$3.2 & 21.1$\pm$4.5 & 27.7$\pm$4.8 & 12.8$\pm$3.2 & 16.9$\pm$5.8 & 7.7$\pm$4.3 \\
 & DAGER & 99.5$\pm$0.5 & 94.0$\pm$4.6 & 87.2$\pm$6.4 & 78.8$\pm$4.7 & 86.4$\pm$5.8 & 71.6$\pm$4.1 & 68.2$\pm$3.2 & 42.5$\pm$2.4 \\
 & \textbf{SOMP}  & \textbf{99.6$\pm$0.4} & \textbf{94.4$\pm$3.6} & \textbf{98.3$\pm$1.7} & \textbf{94.6$\pm$2.3} & \textbf{93.6$\pm$3.9} & \textbf{86.3$\pm$5.5} & \textbf{87.4$\pm$3.4} & \textbf{68.5$\pm$4.5} \\
\hline
 & LAMP  & 73.3$\pm$4.5 & 43.3$\pm$7.0 & 28.6$\pm$3.7 & 11.0$\pm$3.8 & 13.4$\pm$1.4 & 3.9$\pm$1.2 & 8.9$\pm$1.6 & 1.6$\pm$0.8 \\
CoLA & GRAB   & 88.6$\pm$1.6 & 65.7$\pm$1.4 & 84.9$\pm$1.2 & 56.8$\pm$1.2 & 79.2$\pm$0.7 & 52.8$\pm$3.0 & 73.6$\pm$1.7 & 45.6$\pm$5.9 \\
 & DAGER & 100.0$\pm$0.0 & 100.0$\pm$0.0 & 100.0$\pm$0.0 & 100.0$\pm$0.0 & 100.0$\pm$0.0 & 95.6$\pm$1.8 & 99.6$\pm$0.4 & 96.2$\pm$2.1 \\
 & \textbf{SOMP}  & \textbf{100.0$\pm$0.0} & \textbf{100.0$\pm$0.0} & \textbf{100.0$\pm$0.0} & \textbf{98.2$\pm$0.0} & \textbf{99.2$\pm$0.8} & \textbf{94.0$\pm$2.9} & \textbf{100.0$\pm$0.0} & \textbf{97.5$\pm$0.5} \\
\hline
 & LAMP  & 62.2$\pm$6.2 & 31.8$\pm$8.4 & 21.4$\pm$3.2 & 9.5$\pm$3.6 & 9.8$\pm$1.9 & 2.6$\pm$1.4 & 8.1$\pm$0.7 & 0.7$\pm$0.4 \\
SST-2 & GRAB  & 91.2$\pm$0.5 & 63.4$\pm$0.0 & 82.6$\pm$1.7 & 53.7$\pm$0.4 & 69.8$\pm$1.2 & 38.1$\pm$0.6 & 62.7$\pm$0.6 & 33.8$\pm$0.1 \\
 & DAGER & 100.0$\pm$0.0 & 96.0$\pm$3.0 & 100.0$\pm$0.0 & 89.5$\pm$1.4 & 100.0$\pm$0.0 & 92.8$\pm$2.4 & 100$\pm$0.0 & 92.9$\pm$1.6 \\
 & \textbf{SOMP}  & \textbf{100.0$\pm$0.0} & \textbf{98.8$\pm$0.8} & \textbf{100.0$\pm$0.0} & \textbf{97.5$\pm$0.9} & \textbf{100.0$\pm$0.0} & \textbf{98.4$\pm$0.4} & \textbf{99.3$\pm$0.7} & \textbf{95.8$\pm$1.5} \\
\hline
\end{tabular}

\caption{Full reconstruction results for the main experiments, reporting ROUGE-1 and ROUGE-2 across datasets and batch sizes for GPT-J and GPT-2.}
\label{tab:main_results}
\end{table*}

\newpage
\section{Full Main Results}
\label{app:MainExperiment}

This appendix reports the full reconstruction results in Table~\ref{tab:main_results} corresponding to the main experiments in Section~\ref{sec:ExperimentEval}. We include ROUGE-1 and ROUGE-2 across batch sizes $B\in\{1,2,4,8\}$ for all compared methods under the same setup as the main text.

\subsection{Qualitative Reconstruction Example}
\label{ReconstructionExample}
To complement the quantitative results, we present a qualitative reconstruction example under aggregated gradients with batch size $B=2, 4, 8$. Tables~\ref{tab:qual_b2}, \ref{tab:qual_b4}, and \ref{tab:qual_b8} compare the outputs generated by SOMP and several baselines on the same gradient batch. While prior methods such as LAMP, GRAB, and DAGER often produce mismatched or mixed sentences due to cross-sample token interference, SOMP is able to recover the original inputs more accurately.

\section{Multilingual Experiment}
\label{MultilingualExperiment}

This appendix provides additional multilingual reconstruction results in Table~\ref{tab:multilingual-bs} corresponding to Section~\ref{multilingualEval}. We report SOMP's performance across five languages and analyze the effect of tokenization differences on reconstruction quality.

\begin{table}[h]
\centering
\small
\setlength{\tabcolsep}{3pt}
\renewcommand{\arraystretch}{1.05}
\begin{tabular}{lcccc}
\hline
\textbf{Language} &
\textbf{B=1} &
\textbf{B=4} &
\textbf{B=8} &
\textbf{B=16} \\
\hline
English & 99.7$\pm$0.3 & 94.3$\pm$1.4 & 88.2$\pm$1.7 & 83.5$\pm$2.1 \\
German  & 99.6$\pm$0.4 & 94.6$\pm$0.8 & 89.7$\pm$2.1 & 82.3$\pm$2.6 \\
Chinese & 99.8$\pm$0.2 & 93.1$\pm$1.2 & 88.1$\pm$1.4 & 84.4$\pm$3.4 \\
Italian & 99.7$\pm$0.3 & 90.2$\pm$1.5 & 83.3$\pm$1.8 & 78.1$\pm$1.8 \\
French  & 95.7$\pm$2.1 & 86.3$\pm$3.3 & 80.5$\pm$3.6 & 72.8$\pm$3.7 \\
\hline
\end{tabular}
\caption{Multilingual reconstruction performance of \textbf{SOMP} on Qwen3 across batch sizes on \texttt{stanfordNLP/imdb}.}
\label{tab:multilingual-bs}
\end{table}

Across all five evaluated languages, SOMP remains effective without language-specific tuning, indicating robustness to different tokenization schemes. Performance is strongest on English, German, and Chinese, while Italian and French show somewhat larger degradation as batch size increases. This pattern is consistent with heavier subword fragmentation in languages that produce longer tokenized sequences and larger candidate branching during decoding.

\section{Reconstruction under FedAvg}
\label{app:fedavg}

This appendix reports additional results under the FedAvg training protocol, complementing the summary in Section~\ref{sec:fedavg_main}. We evaluate whether multiple local updates before aggregation mitigate gradient leakage.

\paragraph{Experimental Setup.}
We conduct experiments on \texttt{GPT-2} using the Rotten Tomatoes dataset. Unless otherwise specified, we use $E=10$ local epochs, learning rate $\eta = 10^{-4}$, and mini-batch size $B_{\text{mini}} = 4$. Reconstruction quality is measured using ROUGE-1 and ROUGE-2.

\begin{table}[h]
\centering
\small
\setlength{\tabcolsep}{6pt}
\renewcommand{\arraystretch}{1.1}
\begin{tabular}{c c c cc}
\toprule
\textbf{$E$} & \textbf{$\eta$} & \textbf{$B_{\text{mini}}$} &
\textbf{R-1} & \textbf{R-2} \\
\midrule
2  & $10^{-4}$ & 4  &  99.1$\pm$0.9 & 98.2$\pm$1.1 \\
5  & $10^{-4}$ & 4  &  97.5$\pm$1.3 & 95.8$\pm$1.3 \\
10 & $10^{-4}$ & 4  &  95.3$\pm$1.8 & 93.9$\pm$1.6 \\
20 & $10^{-4}$ & 4  &  95.7$\pm$0.9 & 94.4$\pm$1.4 \\
\midrule
10 & $10^{-5}$ & 4  &  99.8$\pm$0.2 & 99.1$\pm$0.9 \\
10 & $5\times10^{-5}$ & 4  &  98.7$\pm$0.9 & 97.9$\pm$1.2 \\
10 & $10^{-4}$ & 4  &  94.5$\pm$1.4 & 93.6$\pm$1.7 \\
\midrule
10 & $10^{-4}$ & 2  &  93.3$\pm$1.9 & 91.4$\pm$2.2 \\
10 & $10^{-4}$ & 8  &  97.7$\pm$1.3 & 96.1$\pm$1.6 \\
10 & $10^{-4}$ & 16 &  99.5$\pm$0.5 & 98.6$\pm$0.8 \\
\bottomrule
\end{tabular}
\caption{Reconstruction performance of SOMP under FedAvg on GPT-2 (Rotten Tomatoes). $E$ denotes the number of local epochs, $\eta$ the learning rate, and $B_{\text{mini}}$ the mini-batch size.}
\label{tab:fedavg}
\end{table}

\paragraph{Results.}
As shown in Table~\ref{tab:fedavg}, FedAvg does not eliminate gradient leakage across a range of reasonable training configurations. SOMP remains capable of recovering sensitive information even when multiple local epochs are performed before aggregation.



\begin{table}[h]
\centering
\small
\setlength{\tabcolsep}{6pt}
\renewcommand{\arraystretch}{1.05}
\begin{tabular}{c cc}
\hline
\textbf{Noise Level} $\boldsymbol{\sigma}$ & \textbf{ROUGE-1} & \textbf{ROUGE-2} \\
\hline
$10^{-5}$           & 78.1 $\pm$ 3.3 & 72.8 $\pm$ 4.1 \\
$5 \times 10^{-5}$  & 55.9 $\pm$ 4.2 & 34.5 $\pm$ 3.9 \\
$10^{-4}$           & 26.7 $\pm$ 4.0 & 16.6 $\pm$ 3.6 \\
$5 \times 10^{-4}$  & 9.6  $\pm$ 3.5 &  1.9 $\pm$ 1.9 \\
\hline
\end{tabular}
\caption{Exploratory reconstruction performance of SOMP under additive Gaussian noise on GPT-2 (Rotten Tomatoes, $B=1$). Higher noise levels correspond to stronger perturbation of the aggregated gradient.}
\label{tab:somp-dp}
\end{table}
\section{Exploratory Results with Differential-Privacy-Inspired Noise}
\label{app:dp}

This appendix provides an exploratory analysis of SOMP under differential privacy--inspired defenses. In particular, we investigate the effect of additive Gaussian noise applied to aggregated gradients. These experiments are preliminary and do not provide formal differential privacy guarantees; they are intended only to assess the qualitative behavior of gradient inversion under noisy training signals.

\paragraph{Experimental Setup.}
Following prior work, we apply zero-mean Gaussian noise with variance $\sigma^2$ to the aggregated gradient $\bm{g}_{\text{mix}}$ before reconstruction. We evaluate SOMP on \texttt{GPT-2} using the Rotten Tomatoes dataset with batch size $B=1$. Reconstruction quality is measured using ROUGE-1 and ROUGE-2.

\begin{table*}[t]
\centering
\small
\renewcommand{\arraystretch}{1.08}
\setlength{\tabcolsep}{3.5pt}
\begin{tabular}{p{0.18\linewidth} p{0.2\linewidth} p{0.2\linewidth} p{0.2\linewidth} p{0.2\linewidth}}
\toprule
\textbf{Reference Inputs} & \textbf{LAMP} & \textbf{GRAB} & \textbf{DAGER} & \textbf{SOMP} \\
\midrule

I received this movie as a gift, I knew from ... movies, don't watch this movie. &
I\hl{camcorder movie"} \hl{DVD sucks filmmaker} \hl{knew wane bee ... movies,} \hl{<br />Sgt this movie.} &
\hl{town watched it pro-} \hl{duce movie, Ben Draper} \hl{... ries HORROR revenge} \hl{woman absolute.} &
I received this movie as a gift, I knew from ... movies, don't watch this movie. &
I received this movie as a gift, I knew from ... movies, don't watch this movie. \\
\cmidrule(lr){1-5}
I very much looked forward to this movie. ... a good love story. &
I very \hl{Michael place} \hl{Willie generally to they ...} \hl{Chris capt. <br /> 's con-} \hl{text 're romantic.} &
\hl{Its Jr.'s editing to much is} \hl{be Too. better care ... love} \hl{Willie and Missy;series.} &
I very much looked forward to\hl{Michael} ... a good love\hl{Michael.} &
I very much looked forward to this movie. ... a good love story. \\
\cmidrule(lr){1-5}
Protocol is an implausible movie whose only ... way to see this film. &
Protocol\hl{Love's implausi-} \hl{ble Missy very much ...} \hl{because see Plus.} &
\hl{Jr.'s is But, implausible} \hl{romantic always only ...} \hl{always enjoy role film.} &
Protocol is an implausible movie whose only ... way to see this\hl{movie.} &
Protocol is an implausible movie whose only ... way to see this film. \\
\cmidrule(lr){1-5}
I was excited to view a Cataluña´s film in ... to go to see this film. &
I was excited\hl{Berlin´s} \hl{a Cataluña´s film in ...} \hl{, was horrible. competition.} &
\hl{Cataluña, too horr to view} \hl{a Cataluña´s see film ...} \hl{to Molina this film.} &
I was excited to view a Cataluña´s film in ... to go to see this film. &
I was excited to view a Cataluña´s film in ... to go\hl{´s} to see \hl{Cataluña´s} film. \\

\bottomrule
\end{tabular}
\caption{
Qualitative reconstruction examples under aggregated gradients with batch size $B=2$. Highlighted spans indicate incorrect tokens caused by cross-sample mixing.
}
\label{tab:qual_b2}
\end{table*}

\paragraph{Results and Discussion.}
As shown in Table~\ref{tab:somp-dp}, reconstruction quality degrades rapidly as the noise magnitude increases. At sufficiently large noise levels, SOMP fails to recover meaningful inputs. A systematic evaluation under formal differential privacy mechanisms, such as DP-SGD with per-sample clipping and privacy accounting, remains an important direction for future work.

\begin{table*}[t]
\centering
\small
\renewcommand{\arraystretch}{1.08}
\setlength{\tabcolsep}{3.5pt}
\begin{tabular}{p{0.18\linewidth} p{0.18\linewidth} p{0.18\linewidth} p{0.18\linewidth} p{0.18\linewidth}}
\toprule
\textbf{Reference Inputs} & \textbf{LAMP} & \textbf{GRAB} & \textbf{DAGER} & \textbf{SOMP} \\
\midrule

A model named Laura is working in South America when ... jungle. It would probably be more fun.<br /><br />3/10 &
A\hl{Laura named ive} \hl{fake Laura in Amer-} \hl{ica America  ... vision.} \hl{far). Franco happening.} \hl{probab be film} fun.<br /><br />3/10 &
\hl{primitives Hunter} \hl{named Devil.<br />} \hl{films working drags} \hl{ity in (25 ...laugh.} \hl{hilariously killed real} \hl{fun more fun.<br /><br} \hl{/>. <br /><br />} &
A model named Laura is working in\hl{hotel} when \hl{Devil Hunter is} \hl{film film her hotel} ... It would probably be more fun.<br /><br />3/10 &
A model named Laura is \hl{work} in South America when ...jungle. It would probably be more fun.3/10\hl{.3/10.3/10.} \\
\cmidrule(lr){1-5}

Devil Hunter gained notoriety for the fact that ... no reason to bother with this one. &
Devil Hunter \hl{film} notoriety for the \hl{'Video} \hl{Nasty' that's films the} \hl{DPP}...\hl{with this one} \hl{devil blood horror.}&
\hl{DPP} Hunter gained notoriety for\hl{films kid-} \hl{napped that title). the} \hl{the ... She Kills in Ec-} \hl{stasy Nasty list but it re-} \hl{ally needn't have been} &
Devil Hunter gained notoriety for the fact that ...\hl{basic Venus film} \hl{films Devil} this one. &
Devil Hunter gained notoriety for the fact that ... no reason to bother with this one\hl{one} \\
\cmidrule(lr){1-5}

This film seemed way too long even ... probably didn't see an uncut version. &
This film \hl{version} \hl{too uncut probably} \hl{way ... scene blood} \hl{camera long didn't} \hl{see}. &
\hl{75 zombie too long} seemed even ... \hl{dub-} \hl{bing didn't rocky} \hl{dubbing way cliff} version. &
This film seemed way too long even ...\hl{film} \hl{films't zombie / flick} \hl{film films...} &
This film seemed way too long even ... probably didn't see an uncut version. \\
\cmidrule(lr){1-5}

I cannot stay indifferent to Lars van Trier's films. I ... to the viewers to decide this alone. &
I\hl{films alone Trier's} \hl{viewers ... cannot} \hl{decide horror camera} \hl{devil Lars stay.} &
\hl{Dancer cannot to} \hl{consider different van} van\hl{Lars's masterpiece} \hl{... pre- ambiguous <br} \hl{/><br /> decide this} \hl{Europa Europa}. &
I cannot stay indifferent to Lars van Trier's films. I ... \hl{differ-} \hl{ent filmmaker van Lars} \hl{van demonstrative this} \hl{alone.'s..} &
I cannot stay\hl{different} to Lars van Trier's films. I ... to the viewers to decide this alone\hl{...} \\
\cmidrule(lr){1-5}

Ah, Channel 5 of local Mexican t.v. Everyday, at 2:00 ... film-making.<br/><br />You have been warned. &
Ah,\hl{warned Mexican} \hl{2:00 Channel ... film-} \hl{making blood devil} \hl{street local been Every-} \hl{day.} &
Ah,\hl{Mexican  Ah, Ah,} \hl{Mexican t.v. Ah,... Ev-} \hl{eryday at 2:00 been} \hl{film-making. You have} \hl{warned.} &
Ah, Channel 5 of local Mexican \hl{Ah, Mexican,} at 2:00...\hl{film-film film.}<br/><br/>\hl{later films,} \hl{cinematography} \hl{warned.} &
Ah, Channel 5 of local Mexican Ah, Everyday, at 2:00 ...\hl{film-} \hl{making.Ah,Ah,}You have been warned.\\

\bottomrule
\end{tabular}
\caption{
Qualitative reconstruction examples under aggregated gradients with batch size $B=4$. Highlighted spans indicate incorrect tokens caused by cross-sample mixing.
}
\label{tab:qual_b4}
\end{table*}

\begin{table*}[t]
\centering
\small
\renewcommand{\arraystretch}{1.08}
\setlength{\tabcolsep}{3.5pt}
\begin{tabular}{p{0.18\linewidth} p{0.18\linewidth} p{0.18\linewidth} p{0.18\linewidth} p{0.18\linewidth}}
\toprule
\textbf{Reference Inputs} & \textbf{LAMP} & \textbf{GRAB} & \textbf{DAGER} & \textbf{SOMP} \\
\midrule

Sloppily directed, witless comedy that supposedly ... be begging for some peace and quiet. (*1/2) &
\hl{peace} \hl{quiet} \hl{comedy} \hl{begging} \hl{sloppily *} ... \hl{hotel} \hl{blood 't} \hl{camera} \hl{witless} \hl{scene}. &
\hl{quiet} \hl{comedy} \hl{peace} ... \hl{witless} \hl{directed 's} \hl{begging} \hl{some} \hl{film} \hl{street}. &
Sloppily directed, witless comedy that supposedly ...\hl{doomed} \hl{James James 50s} \hl{Royale Royale Royale} \hl{Royale..*(*2(.} &
Sloppily directed, witless comedy that\hl{comedy} spoofs ... be begging\hl{some Royale} and \hl{Royale..(*1/2)...} \\
\cmidrule(lr){1-5}

I was disgusted by this movie. No it wasn't ... movie, but find out the true story of Artemisia. &
\hl{movie } disgusted \hl{Artemisia} \hl{true} ... \hl{story} \hl{Beaver} \hl{devil} \hl{blood} \hl{find} \hl{out}. &
\hl{Artemisia} movie \hl{disgusted} ... \hl{Beaver Beaver} \hl{find} \hl{scene} \hl{film}. &
I was disgusted by this movie. No it wasn't ... \hl{<br /><br />} movie, \hl{but movie movie the} \hl{Artemisia Gentileschi.} &
I was disgusted by this movie. No it \hl{it}'t ... movie, but find out the true \hl{Artemisia of Artemi-} \hl{sia....} \\
\cmidrule(lr){1-5}

I didn't think the French could make a bad movie, but ... pornography. So the French can fail, after all! &
\hl{French pornography} \hl{bad} movie ... \hl{after} \hl{fail} \hl{camera} \hl{street} \hl{horror} \hl{French}. &
\hl{pornography}think \hl{French} ... bad movie \hl{fail} after \hl{film} \hl{camera}. &
\hl{I was disgusted by this} \hl{movie. No it wasn't ...} \hl{<br /><br /> movie,} \hl{but movie movie the} \hl{Artemisia Gentileschi.} &
I didn't think the French could make a bad movie, but ... pornography. So the French can fail, after all!\hl{, after all!, after} \hl{all!!!!!} \\
\cmidrule(lr){1-5}

I read somewhere that when Kay Francis refused to ... and you'll be doing yourself a favor. &
\hl{favor} somewhere \hl{yourself} \hl{doing} ... \hl{Kay} \hl{Francis} \hl{camera} \hl{jungle} \hl{blood}. &
\hl{favor}\hl{Francis} somewhere ...\hl{doing yourself} \hl{Kay} \hl{film} \hl{scene}. &
I read somewhere that when Kay Francis refused to ... and \hl{Francis'll movie <br} \hl{/>Also, a favor. And} \hl{her hair!!!}&
I read somewhere that when Kay Francis refused to ... and you'll be \hl{yourself a favor fa-} \hl{vor a a a a...} \\
\cmidrule(lr){1-5}

A stale "misfits-in-the-army" saga, which ... Me", has only two or three brief scenes. What a waste! (*1/2) &
\hl{saga} \hl{brief} \hl{scenes} ... \hl{waste} \hl{army} \hl{devil} \hl{hotel} \hl{camera} \hl{three}. &
\hl{brief films} ... \hl{scenes to} \hl{waste} \hl{army} \hl{film}. &
A stale "misfits-in-the-army" saga, which ... Me", \hl{saga only two} \hl{misfits misfits brief} scenes. What a waste! (*1/2) &
A stale "misfits-in-the-army" saga, which ... Me", has only \hl{two three} scenes. What a waste! (*1/2)! \hl{(*1/2))} \\
\cmidrule(lr){1-5}

When I was at the movie store the other day, I passed up ... and Blonder is a huge blonde BOMBshell.<br /><br />1/10 &
When I \hl{movie store} ... \hl{Michelle} \hl{Romy} \hl{Anderson} \hl{blood} \hl{street} \hl{scene} \hl{day}. &
\hl{Blonder store ... huge} \hl{blonde stupid} \hl{movie Michelle film}. &
When I was at the movie store the other day, I passed up ... and Blonder is a huge  \hl{BOMB.}<br /><br />1/10\hl{<br /><br />1/10} &
When I was at the movie store the other day, I passed up ... and Blonder is a  blonde \hl{Blonder}.<br /><br />1/10\hl{.....}D \\

\bottomrule
\end{tabular}
\caption{
Qualitative reconstruction examples under aggregated gradients with batch size $B=8$. Highlighted spans indicate incorrect tokens caused by cross-sample mixing.
}
\label{tab:qual_b8}
\end{table*}

\end{document}